\documentclass[sigconf]{acmart}
\usepackage{multirow}
\begin{document}

\title{A Multi-Modal Knowledge-Enhanced  Framework for Vessel Trajectory Prediction}

\author{Haomin Yu, Tianyi Li$^{*}$, Kristian Torp, Christian S. Jensen }
\email{{haominyu, tianyi, torp, csj}@cs.aau.dk}
\affiliation{%
  \institution{Aalborg University}
  \city{Aalborg}
  \country{Denmark}
}

\renewcommand{\shortauthors}{Haomin Yu et al.}

\begin{abstract}
Accurate vessel trajectory prediction facilitates improved navigational safety, routing, and environmental protection. However, existing prediction methods are challenged by the irregular sampling time intervals of the vessel tracking data from the global AIS system and the complexity of vessel movement. These aspects render model learning and generalization difficult. To address these challenges and improve vessel trajectory prediction,  we propose \textbf{M}ulti-mod\textbf{A}l \textbf{K}nowledge-\textbf{E}nhanced f\textbf{R}amework (\textsf{MAKER})  for vessel trajectory prediction. To contend better with the irregular sampling time intervals, \textsf{MAKER} features a Large language model-guided Knowledge Transfer (LKT) module that leverages pre-trained language models to transfer trajectory-specific contextual knowledge effectively. To enhance the ability to learn complex trajectory patterns, \textsf{MAKER} incorporates a Knowledge-based Self-paced Learning (KSL) module. This module employs kinematic knowledge to progressively integrate complex patterns during training, allowing for adaptive learning and enhanced generalization. Experimental results on two vessel trajectory datasets show that \textsf{MAKER} can improve the prediction accuracy of state-of-the-art methods by 12.08\%---17.86\%.
\end{abstract}

\maketitle

\begin{figure}[!h]
    \centering
    \includegraphics[width=0.75\linewidth]{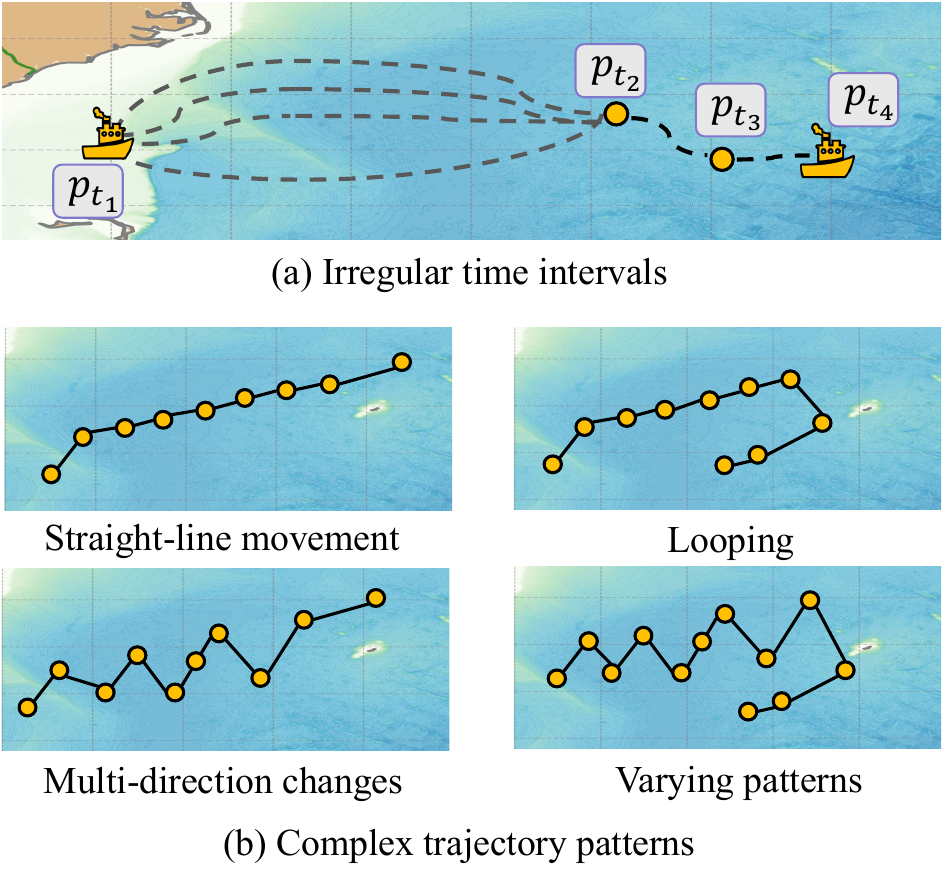}
   \caption{Challenges to vessel trajectory prediction.}
   \label{fig:intro}
\end{figure}
\section{Introduction}
A cornerstone of the global economy, maritime transportation enables the efficient and cost-effective distribution of commodities worldwide~\cite{koccak2021comparative}. Vessel trajectory prediction plays a critical role in enhancing maritime transportation by enabling improved navigational safety and optimizing routes. Traditionally, vessel trajectory tracking relies heavily on satellite communications, which comes with high data transmission and infrastructure maintenance costs~\cite{cervera2011satellite}.  Recent advancements in data mining and time series analysis have improved vessel trajectory prediction~\cite{qi2016trajectory,tang2022model,forti2020prediction}. These data mining techniques enhance data processing and predictive capabilities, while time series methods provide competitive performance by exploiting trend dependencies in trajectory data.

%To enhance vessel trajectory prediction, advanced time series methods~\cite{tang2022model,forti2020prediction} are commonly employed to achieve competitive accuracy by extracting and analyzing trend dependencies.

% The recent, rapid developments in data mining have attracted tremendous interest as a means of facilitating vessel trajectory prediction~\cite{qi2016trajectory}. 

%Given its role in handling large volumes of bulk commodities, such as oil and minerals, analyzing maritime traffic is essential for optimizing globalBen Trovato trade logistics~\cite{zhou2021classification}. 

%Traditional methods for determining vessel trajectory often rely on satellite communication systems~\cite{cervera2011satellite}, which involve significant expenses due to the high costs associated with satellite data transmission and maintenance of the necessary infrastructure. The recent, rapid developments in data mining have attracted tremendous interest as a means of facilitating vessel trajectory prediction~\cite{qi2016trajectory}. To enhance vessel trajectory prediction, advanced time series methods~\cite{tang2022model,forti2020prediction} are commonly employed to achieve competitive accuracy by extracting and analyzing trend dependencies.

However, vessel trajectory prediction encounters several key challenges.
\textit{First, the data points in AIS data trajectories are sampled at irregular intervals due to varying environmental conditions, which causes varying and large time gaps between temporally consecutive data points.}  Fig.~\ref{fig:intro}(a) shows four trajectory points $g_{t_i}\,(1\leq i \leq 4)$, with a larger time gap between $g_{t_1}$ and $g_{t_2}$ than between the next points. The irregular sampling intervals in AIS data complicate the use of machine learning models, which typically expect more regular data. A naive strategy would be to apply interpolation to create uniformly sampled data. However, this interpolation strategy can introduce significant errors and may not capture the actual dynamics, particularly when there are large time interval gaps in the data.

\textit{Second, vessel trajectories are shaped by numerous factors, such as weather conditions and human decisions, causing trajectories to capture complex behaviors with hard-to-discern patterns.} Fig.~\ref{fig:intro}(b) depicts four distinct movement patterns: straight-line movement (e.g., long-distance travel to minimize fuel consumption), looping (e.g., fishing),  multiple direction changes (e.g., traveling in busy waters to avoid collisions), and the varying patterns (e.g., mixed movements in complex environments). This illustrates that movement patterns can vary over time involving multiple direction changes and looping. It is a challenge to capture and exploit such complex patterns for prediction. Complex trajectory patterns can significantly hinder a model’s capacity to generalize effectively across diverse data distributions.

To address the above challenges, we propose the Multi-modAl Knowledge-Enhanced fRamework~(\textsf{MAKER}) for vessel trajectory prediction. First, to alleviate the influence of irregular time intervals, we develop a large language model (LLM)-guided knowledge transfer (LKT) module that combines contextual knowledge with pre-trained LLMs using predefined textual prompts. This module utilizes a multi-modal framework, combining a masked sequence encoder with an LLM-guided sequence encoder to integrate trajectory records and textual prompts. Considering the gaps between multi-modal data, the LKT module also incorporates a multi-modal knowledge transfer component to effectively bridge these gaps, facilitating seamless interaction between diverse data modalities.
Second, to alleviate the influence of complex trajectory patterns, we employ self-paced learning with kinematic knowledge guidance. This learning approach begins with easier trajectories and progressively includes more challenging ones. By gradually increasing the difficulty of the samples, the model can better capture intricate patterns and relationships in the data. 

Our contributions are summarized as follows.

%Guided by kinematic knowledge such as speed and acceleration, this approach enables the model to build a strong foundation before tackling more complex trajectories.

\begin{itemize}
    \item  We propose a novel multi-modal knowledge-enhanced framework for accurate vessel trajectory prediction called \textsf{MAKER} to enhance trajectory prediction by integrating multi-modal information. %\textsf{MAKER} enhances trajectory prediction with greater informativeness and contextual depth.
    
    \item \textsf{MAKER} includes an LLM-guided Knowledge Transfer (LKT) module, leveraging extensive pre-trained knowledge and contextual understanding from predefined textual prompts to effectively interpret and process irregularly sampled trajectory points.

    \item \textsf{MAKER} includes a Knowledge-based Self-Paced Learning (KSL) module that incrementally integrates complex patterns into the training process. This enables an adaptive transition from simple to more complex patterns, guided by kinematic knowledge to improve the handling of complex trajectory data.

    \item  We report on experiments on two real vessel trajectory datasets, finding that \textsf{MAKER} can outperform seven state-of-the-art methods by 12.08\%---17.86\%.
\end{itemize}
%\cite{tang2022model} \cite{forti2020prediction} \cite{capobianco2021deep}

\section{Related Work}
\subsection{Vessel Trajectory Prediction}
Vessel trajectory prediction has benefited increasingly from the application of machine learning techniques. Traditional methods that rely mainly on statistical approaches~\cite{mazzarella2015knowledge,alizadeh2021prediction} often face limitations in capturing complex patterns inherent in vessel movement. As a result, there is a growing research interest in leveraging deep learning methods, which offer enhanced capabilities for modeling the complex dynamics of vessel trajectories.

% Long Short-Term Memory (LSTM) model is utilized~\cite{tang2022model} to predict vessel trajectories. However, LSTMs face limitations in handling variable-length input and output sequences.
Forti et al.~\cite{forti2020prediction} propose a sequence-to-sequence model based on LSTMs to effectively capture long-term temporal dependencies. However, the sequence-to-sequence model struggles to maintain long-term dependencies, causing reduced effectiveness for vessel trajectory prediction tasks that require the capture of long-term trend dependencies. Capobianco et al.~\cite{capobianco2021deep} propose an attention-based sequence-to-sequence architecture for trajectory prediction, utilizing an attention mechanism to enhance the focus on relevant long-term temporal dependencies. However, despite the incorporation of attention, the model still struggles to effectively capture complex patterns in trajectory data recorded at irregular intervals.

Unlike previous studies that focus on the integration of sequence prediction models for trajectory prediction, \textsf{MAKER} 
incorporates LLMs to learn
both trajectory records and predefined textual prompts, thereby enhancing trajectory prediction through the integration of contextual knowledge.

  \begin{figure*}
    \centering
    \includegraphics[width=0.88\linewidth]{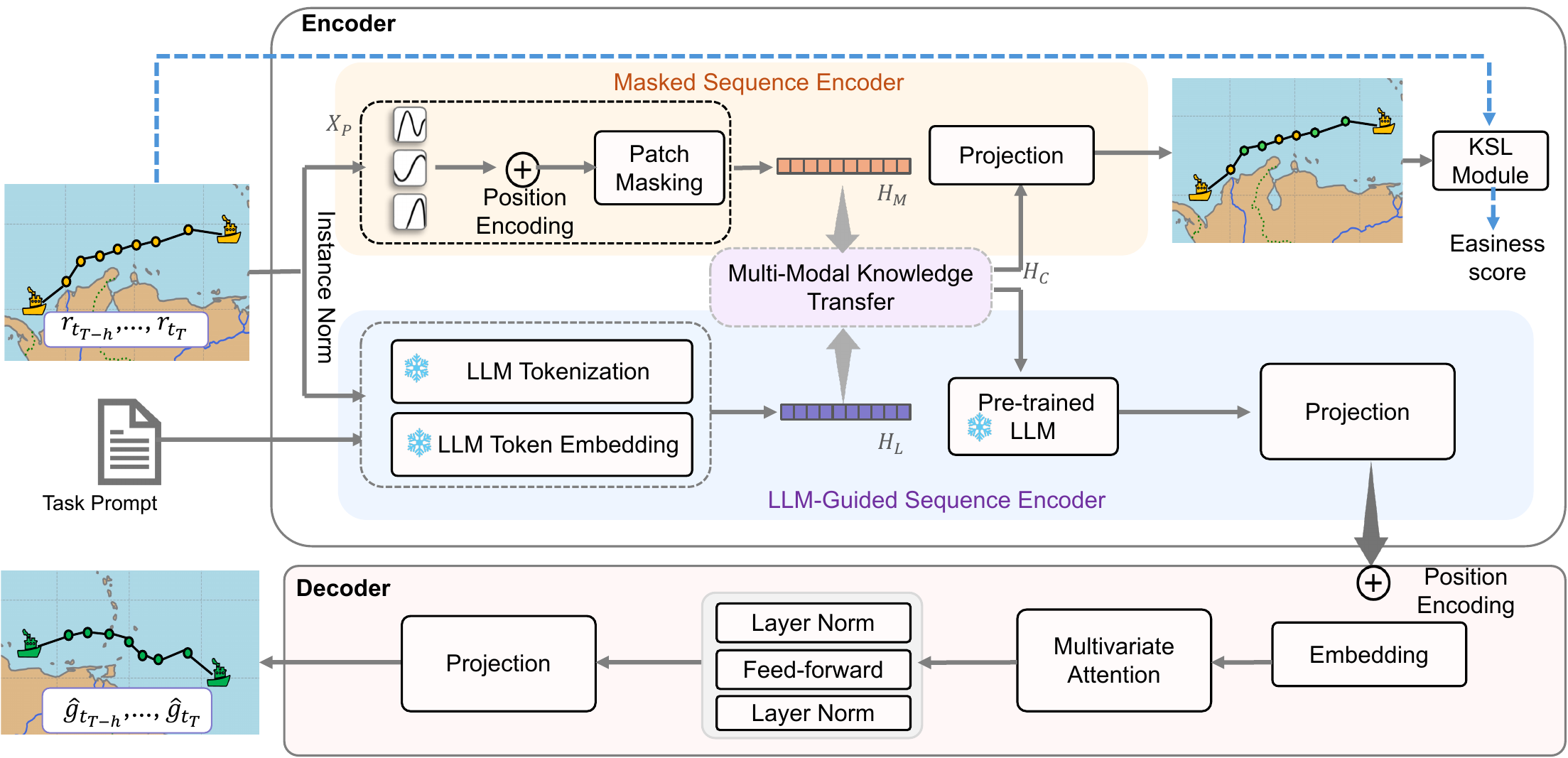}
    \caption{Multi-modal knowledge-enhanced framework for vessel trajectory prediction.}
    \label{fig:pip}
\end{figure*}

\subsection{Self-paced Learning}
Curriculum Learning~\cite{bengio2009curriculum}, inspired by the effective learning progression in humans and animals, emphasizes mastering simpler samples before gradually progressing to more complex ones, thereby enhancing both model convergence and generalization.
However, this approach relies on the ability to determine the easiness of samples, which can be challenging. Thus, self-paced learning (SPL)~\cite{kumar2010self}  is introduced as a method that dynamically determines sample easiness and adapts its learning process by following an easy-to-hard paradigm based on a predefined criterion (e.g., loss value or confidence). The number of samples included in the training is controlled by a gradually annealed weight, progressing until the entire training dataset is utilized.

SPL has been applied widely across different domains, including image classification~\cite{xu2018multi,wu2017semi}, natural language processing~\cite{raamadhurai2019curio,wan2020self}, and complex fields such as medical diagnosis~\cite{wang2022fedspl,wang2020adaptive} and time series analysis~\cite{yu2023cgf,li2023self,tang2019tensor,yang2024self}. Its success in these domains, particularly in the time series domain, highlights its potential for advancing the analysis of vessel trajectory data. 

Given the complexity of trajectory patterns, we incorporate kinematic knowledge, such as speed and acceleration, into our SPL method to determine sample easiness. By gradually increasing the difficulty of samples, this approach aims to enable models to progressively capture intricate patterns in the trajectory data, strengthening their ability to contend with complex trajectories.

\section{Preliminaries}
\subsection{Definitions}

\noindent \textbf{Definition 1 (Trajectory)}.
A vessel trajectory is a sequence of points  $( {o}_{t_1}, {o}_{t_2}, \ldots, {o}_{t_{T}})$, where each point ${o}_{t_i} = (g_{t_i}, t_i) $ and $g_{t_i}=(\textit{lon}_{t_i}, \textit{lat}_{t_i})$.  Here, $\textit{lon}_{t_i}$ and $\textit{lat}_{t_i}$ are a longitude and a latitude.

% ${g}_{t_i}= (g_{t_i}, t_i)$  and 

\subsection{Problem Formulation}
We focus on predicting the future trajectory of marine vessels. Formally, let the current time be $t_T$. Our goal is to use the past $h$ historical trajectory records to predict the vessel's trajectory for the next $p$ timestamps. %Considering that the time intervals between consecutive steps are irregular, we need to account for these variations in our model.

The input to the model consists of $h$ historical records denoted as $({r}_{t_{T-h}}, {r}_{t_{T-h+1}}, \ldots, {r}_{t_T}) \in \mathbb{R}^{h \times (M+2)}$, where each $r_{t_i}$ represents record data at a timestamp $t_i$. $r_{t_i}$ includes $g_{t_i} \in \mathbb{R}^{2}$ and other available features ${q}_{t_i} \in \mathbb{R}^{M}$, where $M$ is the number of features.  The output is the future trajectory $(\hat{g}_{t_{T+1}}, \hat {g}_{t_{T+2}}, \ldots, \hat {g}_{t_{T+p}}) $, where $\hat{g}_{t_{T+i}} $ represents the predicted trajectory at timestamp $t_{T+i}$.

 Specifically, the objective is to build a prediction model $f$ that, given the historical trajectory $(g_{t_i})_{{i=T-h}}^{T}$, other available data  $({q}_{t_i})_{i=T-h}^{T}$ and historical timestamps  $(t_i)_{i=T-h}^{T}$, and the future timestamps  $(t_i)_{i=T+1}^{T+p}$, accurately predicts the future trajectory:
%that need to predict into  model, and then accurately predicts the future $\{\hat{{{g}}}\}_{t_{T+1}}^{t_{T+P}}$ time steps of the trajectory:
\begin{equation}
   (\hat{{{g}}}_{t_i})_{i={T+1}}^{{T+p}} =f((r_{t_i})_{i={T-h}}^{T},   (t_i)_{i=T-h}^{T}, (t_i)_{i=T+1}^{T+p})
\end{equation}
The goal of the prediction model $f$ is to make the predicted trajectory 
$(\hat{g}_{t_{T+1}}, \hat{g}_{t_{T+2}}, \ldots, \hat{g}_{t_{T+p}}) $ approximate the actual future trajectory $(g_{t_{T+1}}, g_{t_{T+2}}, \ldots, g_{t_{T+p}})$ closely.

% $\{\hat{g}_{t_{T-h}},  \ldots, \hat{g}_{t_{T}}\} $
\section{Method}

\subsection{Framework Overview}
We propose a multi-modal knowledge-enhanced framework to enable models the capture of complex patterns in irregularly sampled trajectories. As illustrated in Fig.~\ref{fig:pip}, the proposed framework comprises two main stages: an encoding stage and a decoding stage.

In the encoding stage, we aim to learn trajectory trend patterns from historical records, i.e., $(r_{t_i})_{{i=T-h}}^{T}$, supported by predefined textual prompts, by leveraging two key modules: the LKT module and the KSL module. 

The LKT module consists of three components. As illustrated in the yellow section in Fig.~\ref{fig:pip}, the masked sequence encoder starts by applying instance normalization to the historical records $(r_{t_i})_{{i=T-h}}^{T}$, resulting in the normalized trajectory representation ${X}_p$.
This representation is then divided into $Q$ patches of length $L_p$, which are processed through patch encoding and combined with position encoding to form ${H}_P$. A masking strategy is applied to obtain the representation ${H}_M$, which is ultimately used for reconstructing the historical trajectory positions $({\hat{g}}_{t_i})_{i={T-h}}^{T}$ through a reconstruction projection.
As shown in the purple section in Fig.~\ref{fig:pip}, the LLM-guided sequence encoder integrates contextual knowledge from textual prompts into the trajectory analysis. It combines the normalized trajectory $(g_{t_i})_{i={T-h}}^{T}$ with a textual prompt, and the LLM tokenizes and embeds this combined input, generating the contextual trajectory representation ${H}_L$.
Moreover, as shown in the green part in Fig.~\ref{fig:pip}, the multi-modal knowledge transfer component combines the sequence representation ${H}_M$ and the LLM-derived representation ${H}_L$, bridging the gap between the two modalities to form a unified representation.

The KSL module complements the LKT module by leveraging kinematic knowledge to guide the training process. Through self-paced learning, the KSL module determines whether a sample should be included in the training sequence, thereby optimizing the learning process and enhancing overall training effectiveness.

In the decoding stage,  we process the representation generated by the encoder to predict future trajectories. Specifically, we pass the encoded representation ${H}_G$ to iTransformer~\cite{liu2023itransformer} to get the final prediction results $({{\hat{g}}_{t_i}})_{i={T+1}}^{{T+p}}$.

We proceed to present the two key modules in detail.

\subsection{LLM-Guided Knowledge Transfer Module}

\subsubsection{Motivation}
Inspired by the capability of LLMs, which are pre-trained on extensive corpora to efficiently process complex patterns across various sequence analysis tasks, we integrate LLMs into our trajectory prediction framework. However, since LLMs are generally optimized for generalized tasks, their direct applicability to domain-specific scenarios like trajectory prediction may be limited. To address this, we incorporate a dedicated sequence encoder specifically designed for recognizing sequence patterns in trajectory records. This sequence encoder encodes trajectory data while an LLM-guided sequence model learns domain-specific knowledge using predefined textual prompts. To facilitate the transfer of knowledge between distinct modalities, i.e., trajectory records and textual data, we introduce a multi-modal knowledge transfer component to integrate insights from multiple perspectives.

The following sections provide details of the masked sequence encoder, the LLM-guided sequence model, and the multi-modal knowledge transfer component.
\begin{figure}
    \centering
    \includegraphics[width=0.8\linewidth]{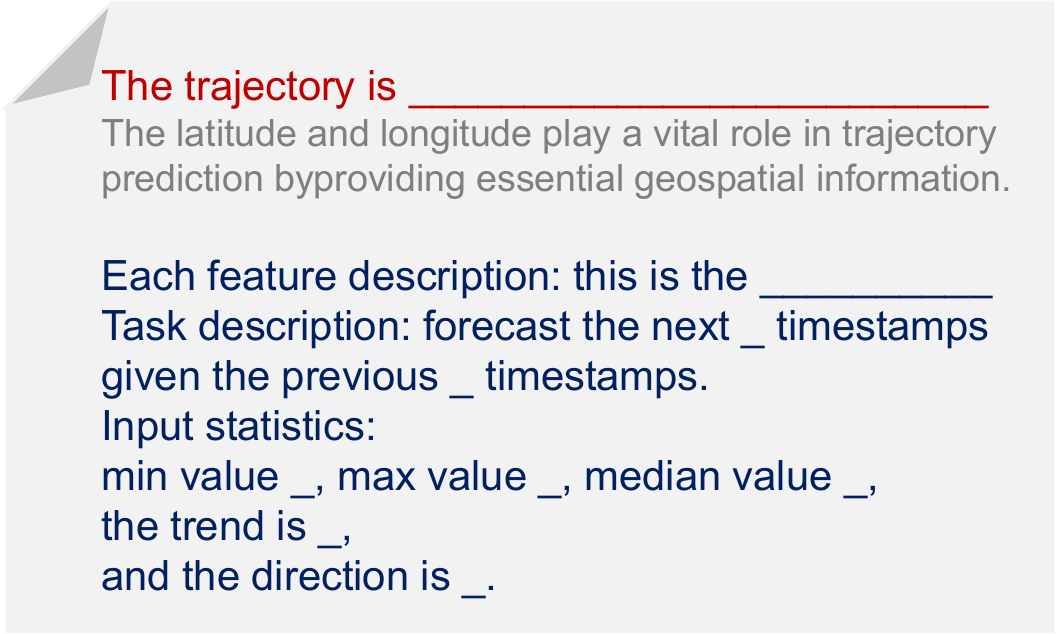}
    \caption{The textual prompt, with underlined spaces indicating where sample-specific data is inserted.}
    \label{fig:prompt}
\end{figure}

\subsubsection{Masked Sequence Encoder}
A masked sequence encoder focuses on extracting patterns from historical records to learn the pattern regularities for better understanding and predicting future trajectories based on past behavior. 

First, we process the historical record data $(r_{t_i})_{{i=T-h}}^{T} \in \mathbb{R}^{h \times (M+2)}$ using instance normalization, resulting in ${X}_P \in \mathbb{R}^{h \times (M+2)}$. This multivariate time series is decomposed into $M+2$ univariate time series, which are independently processed. The sequences are subsequently divided into patches of length $L_p$, segmenting ${X}_P$ into multiple patches, denoted as ${X_P'}$. Each channel $X_P^{\prime(i)}$ is represented in $\mathbb{R}^{Q \times L_p}$, where $Q$, the total number of input patches, is calculated as $Q = \left\lfloor \frac{(h - L_p)}{S} \right\rfloor + 2$, with $h$ representing the sequence length, $L_p$ the patch length, and $S$ the stride.

These patches are then embedded into ${H}_P$, where each channel $ {H}_P^{(i)} \in \mathbb{R}^{P \times d_m}$. Simultaneously, the corresponding timestamps $(t_i)_{i=T-h}^{T}$ are embedded into vectors of size $\mathbb{R}^{P \times d_m}$ using timestamp embeddings. The two embeddings are combined via an ``add'' operation to obtain ${H}_M$. A masking technique~\cite{lee2023learning} is applied to each channel of ${H}_M$ to mask half of the patches, producing ${H}_M^{(i)} \in \mathbb{R}^{P \times d_m}$. Finally, a reconstruction projection is used to reconstruct the trajectory $(g_{t_i})_{i={T-h}}^{T}$.

The masked sequence encoder is designed to reconstruct the masked portions of the trajectory record, requiring the model to leverage contextual information from the unmasked segments. This process encourages the model to develop a more comprehensive understanding of the trend patterns within the historical trajectory data.

\subsubsection{LLM-guided Sequence Model}
The masked sequence encoder excels at processing historical records and capturing trajectory trends over time. However, irregular collection intervals complicate the task of accurately modeling historical trajectory patterns. LLM, as pre-trained models with extensive knowledge, excel in semantic understanding by effectively leveraging predefined textual prompts. We can leverage these capabilities by integrating LLM with temporal models~\cite{gruver2024large,jin2023time,yu2023temporal} to enhance the temporal comprehension of contextual information. Then, we detail the LLM-guided sequence model.

First, we convert the historical trajectory information into a string format, while simultaneously setting a textual prompt to inform the model of the task, along with the trajectory and other relevant statistical information, all represented as a string ${S}$, as shown in Fig.~\ref{fig:prompt}.
It is then tokenized into a sequence of tokens  ${T_s} = \mathrm{Tokenize}({S})$, allowing us to process the data in a form suitable for embedding.
The token sequence ${T_s}$ is then mapped to an embedding space, resulting in token embeddings ${H}_L$, expressed as ${H}_L = {E}({T_s})$, where ${E}$ is the embedding function that converts each token into a high-dimensional vector, capturing semantic meanings.

\begin{figure}
    \centering
    \includegraphics[width=\linewidth]{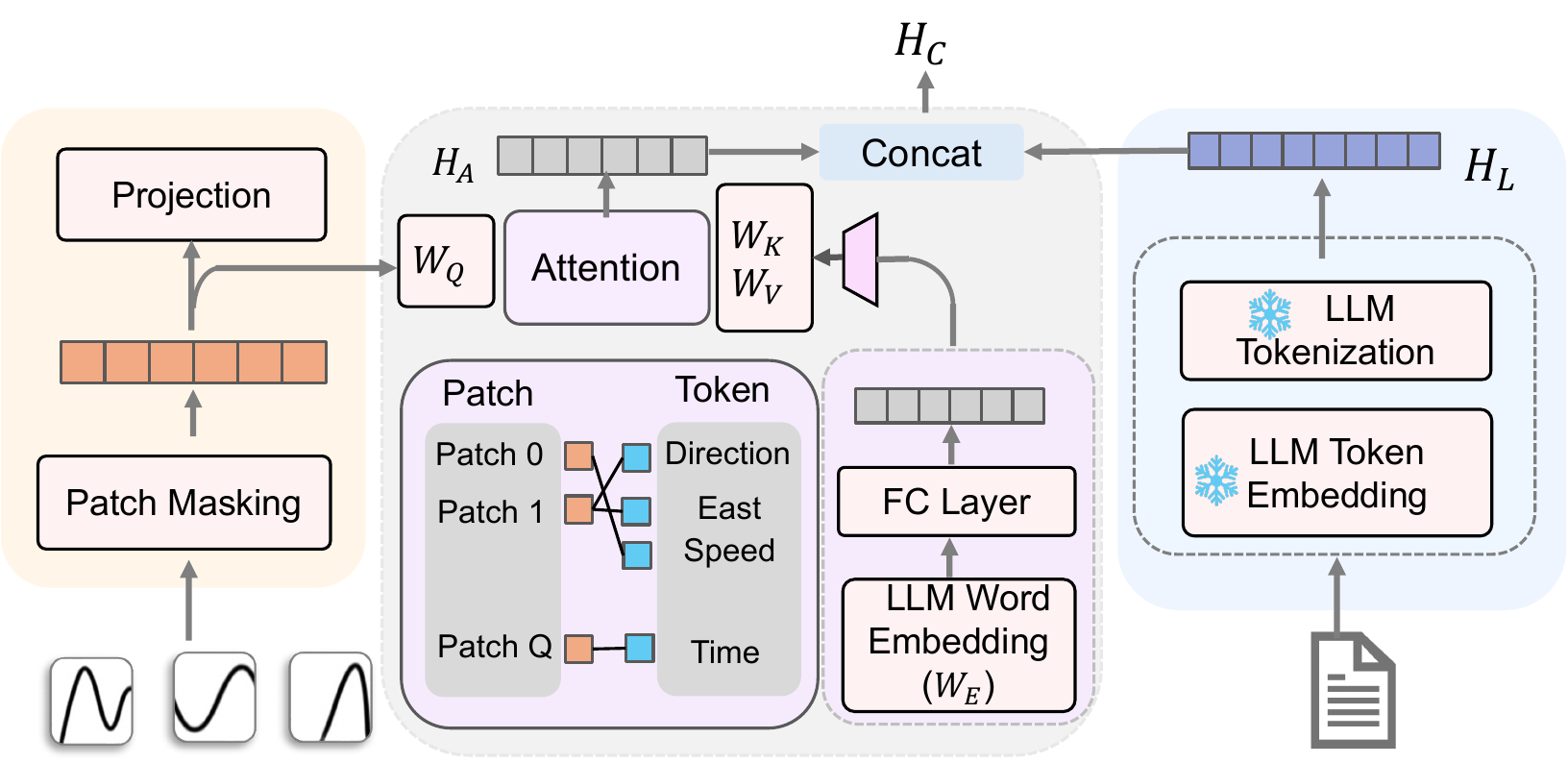}
    \caption{Multi-modal knowledge transfer component.}
    \label{fig:com}
  
\end{figure}

\subsubsection{Multi-modal Knowledge Transfer Component}
While LLMs handle textual prompts, the masked sequence encoder processes trajectory sequential data. In this section, we introduce a multi-modal knowledge transfer module to bridge gaps between two different modalities.  This component learns the correlations between different data representations through hidden layers, facilitating the fusion of multiple data modalities.

Specifically, we utilize the pre-trained word embedding ${W}_E$ to guide the knowledge transfer procedure, which aims to establish a representational bridge between the trajectory sequence modality and the textual modality, as shown in Fig.~\ref{fig:com}. The pre-trained word embedding ${W}_E$ is first sent to a linear layer and obtains a word representation ${H}_E$.

Then, the attention mechanism enables the model to selectively identify relevant parts between textual and sequential modalities, effectively bridging gaps between them by aligning their representations and enhancing cross-modal understanding. We use the sequence representation ${H}_M$ to construct the query matrices ${Q}^{(i)} = {{H}}_M^{(i)} {W}_Q$. This choice is driven by the need for the model to query relevant contextual information from the textual modality that can help interpret and enhance the understanding of trajectory patterns. The textual modality, embedded through the LLM, is used as both the key and value in the attention mechanism, with key matrices ${K}^{(i)} = {H}_E {W}_K$ and value matrices ${V}^{(i)} = {H}_E {W}_V$. This setup leverages the rich semantic knowledge in the textual data to guide and refine the trajectory sequence analysis.

By setting the trajectory sequence as the query, we enable the model to actively seek out relevant information from the textual data, effectively using the LLM to contextualize and interpret the sequence data. The dimensions of the matrices are defined as ${W}_Q \in \mathbb{R}^{d_m \times d_m}$ and ${W}_K, {W}_V \in \mathbb{R}^{D \times d_m}$, where $D$ denotes the hidden dimension.  The attention mechanism~\cite{vaswani2017attention} is defined as follows:

\begin{equation}
    \begin{aligned}
    {H}_A^{(i)} &= \text{ATTENTION}({Q}^{(i)}, {K}^{(i)}, {V}^{(i)}) \\
    &= \text{SOFTMAX}\left(\frac{{Q}^{(i)} {K}^{(i)\top}}{\sqrt{d}}\right) {V}^{(i)}.
    \end{aligned}
\end{equation}
Having obtained the  ${H}_A^{(i)} \in \mathbb{R}^{P \times D}$, 
we concatenate  it with the LLM representation ${H}_L$ to obtain ${H}_C$.

\subsection{Knowledge-based Self-paced Learning (KSL) Module}
\subsubsection{Motivation}
The Knowledge-based Self-paced Learning (KSL) module leverages domain knowledge to guide the learning process, allowing the model to prioritize easier and more informative samples during the initial training stages and gradually address more complex patterns. This approach ensures more effective learning by enabling the model to build a solid foundation before tackling challenging trajectory patterns. However, a critical aspect of training with self-paced learning is determining how to effectively assess the easiness of samples. To address this, we integrate domain-specific insights, particularly kinematic knowledge, to guide the model in better understanding the easiness of trajectory data. This knowledge-guided progression not only enhances the model’s ability to manage data complexity but also improves overall learning efficiency.

\begin{figure}
    \centering
    \includegraphics[width=0.85\linewidth]{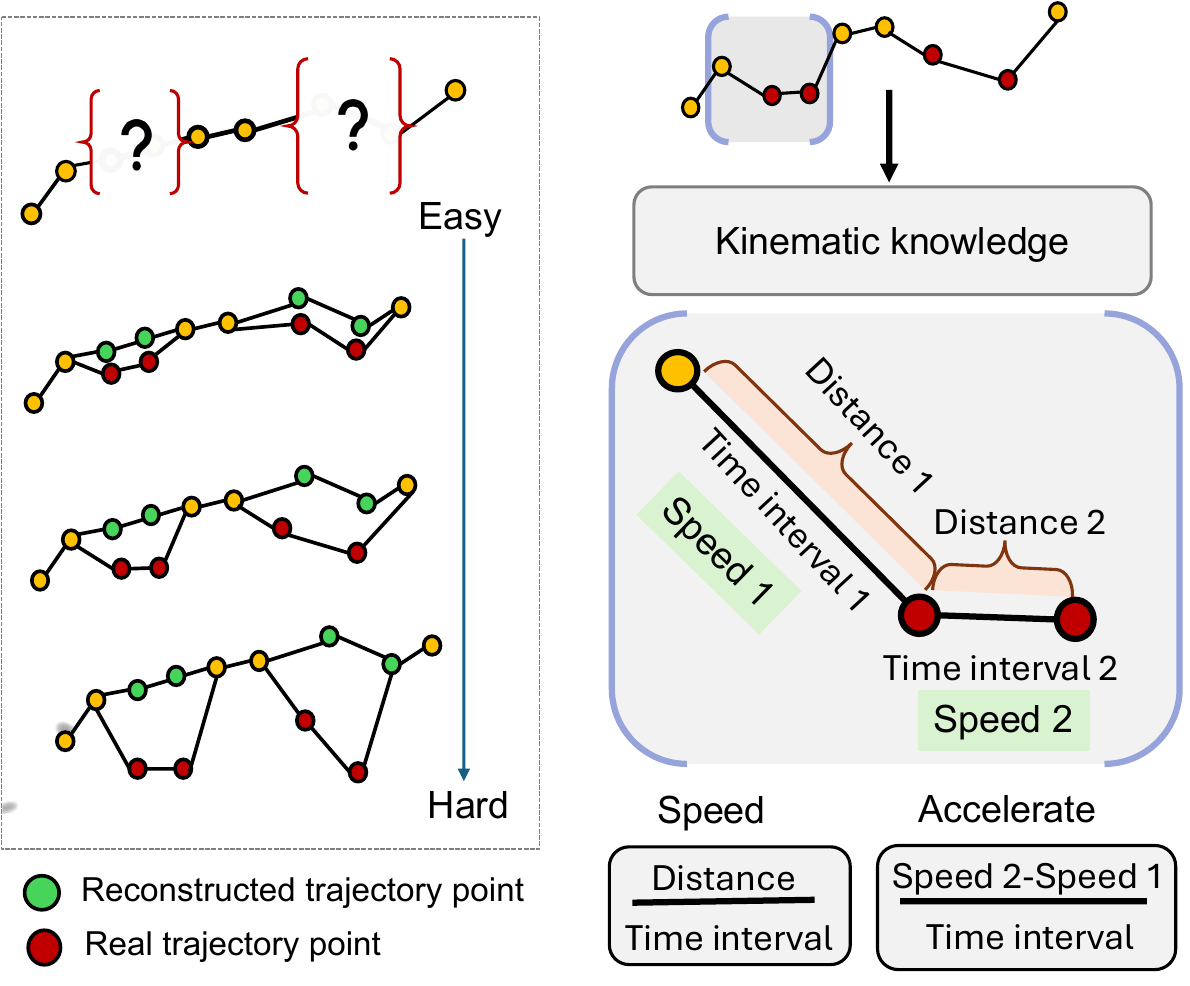}
    \caption{Knowledge-based self-paced learning (KSL) module}
    \label{fig:spl-label}
\end{figure}
\subsubsection{Self-Paced Learning Component}
Inspired by self-paced learning (SPL), we designed a component that leverages domain-specific knowledge to gradually learn more complex trajectories, allowing the model to effectively build on simpler patterns before tackling greater complexities, as shown in Fig.~\ref{fig:spl-label}.

For convenience, we first define the historical trajectory $\{g_{t_i}\}_{i={T-h}}^{T}$ as ${Y}_i$, the historical record $\{r_{t_i}\}_{i={T-h}}^{T}$ as ${X}_i$. %, and the reconstruction data $\{{\hat{g}}_i\}_{i=t_{T-h}}^{t_T}$ as ${\hat{Y}}_i$.
The SPL primarily focuses on the trajectory reconstruction process. Given the mini-batch training dataset $\mathcal{D} = \{({X}_1, {Y}_1), \ldots, ({X}_n, {Y}_n)\}$, let $L({Y}_i, f_r({X}_i, {w}))$ represent the loss function that computes the loss between the ground truth label ${Y}_i$ and the predicted label $f_r({X}_i, {w})$. Here, ${w}$ represents the model parameters within the decision function $f_r$. In SPL, the goal is to jointly learn the model parameters ${w}$ and the latent weight variable ${v} = [v_1, \ldots, v_n]$ by minimizing:
\begin{equation}
  E(w, v; \lambda) = \sum_{i=1}^n v_i L({Y}_i, f_r({X}_i, {w})) - \lambda \sum_{i=1}^n v_i, \quad {v} \in [0, 1]^n,
\end{equation}
where $\lambda$ is a parameter that controls the learning pace. 
Alternative Convex Search (ACS)~\cite{kumar2010self} is generally used to solve this optimization problem. ACS is an iterative method for biconvex optimization, where the variables are divided into two disjoint blocks. In each iteration, one block of variables is optimized while keeping the other block fixed. When ${v}$ is fixed, existing supervised learning methods can be employed to obtain the optimal ${w}^*$. With ${w}$ fixed, the global optimum ${v}^* = [v_1^*, \ldots, v_n^*]$ can be easily calculated by:
\begin{equation}
v_i^* = \begin{cases} 1 & \text{if } L({Y}_i, f_r({X}_i, {w})) < \lambda, \\ 0 & \text{otherwise}. \end{cases}
\end{equation}
Here, ${v}$ represents the easiness score, which determines whether a trajectory is difficult to forecast. The easiness score is a Boolean value that controls the input of a forecasting unit. If the easiness score of a trajectory is 1, the trajectory is considered to be an easy sample and is used as input to the model training. Otherwise, it is discarded. When updating ${v}$ with a fixed ${w}$, a sample whose loss is smaller than a certain threshold $\lambda$ is considered an "easy" sample and is selected for training $(v_i^* = 1)$; otherwise, it is unselected $(v_i^* = 0)$.
When updating ${w}$ with a fixed ${v}$, the model is trained only on the selected  ``easy'' samples.
The parameter $\lambda$ controls the pace at which the model learns new samples. When $\lambda$ is small, only ``easy'' samples with small losses are considered. As $\lambda$ increases, more samples with larger losses are gradually included to train a more ``mature'' model.

% \usepackage{multirow}

\subsubsection{Kinematic Knowledge Guidance Component}
The design of the loss function $L(\cdot)$ is crucial in self-paced learning because it affects directly how the model evaluates and prioritizes samples during training. A well-designed loss function ensures that easier samples with lower loss values are selected first, allowing the model to learn progressively.

The kinematic knowledge guidance introduces a novel approach to enhance the loss function $L(\cdot)$, i.e., the mean absolute error,  by integrating predictions of velocity and acceleration. The ground truth of velocity is calculated as the distance between points over time, and acceleration is calculated by the change in velocity over time between intervals.
By tasking the model with predicting the trajectory and estimating the corresponding velocity and acceleration, we introduce a kinematic loss constraint to guide the learning process.
This dual focus allows the model to gain a deeper understanding of motion dynamics, improving its ability to capture intricate patterns and dependencies in the data. The kinematic loss helps ensure that the model’s predictions are physically plausible and consistent with real-world movement behaviors. This approach enhances predictive accuracy and robustness, making it particularly effective for complex trajectory prediction tasks. By leveraging Kinematic Knowledge Guidance within the loss function $L(\cdot)$, the model is explicitly encouraged to predict physically consistent velocity and acceleration alongside trajectory predictions. 
The effectiveness of the kinematic guidance is evaluated in Section~\ref{abstud}.

\begin{table}[!h]
\caption{Dataset description}

\centering
\renewcommand{\arraystretch}{1.2}
 \smallskip\resizebox{\columnwidth}{!}{
\begin{tabular}{c|c|c|c|c}
\toprule[1.2pt]

\multirow{2}{*}{Dataset} & \multirow{2}{*}{Start\_time} & \multirow{2}{*}{End\_time} & \multirow{2}{*}{\#Sample} & \multirow{2}{*}{\begin{tabular}[c]{@{}c@{}} Interval\end{tabular}} \\
                         &                              &                            &                           &                                                                              \\ \hline
US Coast                 & 2023-12-25                   & 2023-12-31                 & 50,961,322                & 3 min                                                                        \\ \hline
Danish Maritime          & 2024-02-16                   & 2024-02-22                 & 108,698,248               & 1 min                                                                        \\  \toprule[1.2pt]
\end{tabular}}
\label{daes}
\end{table}
\begin{table*}
\centering

\caption{Forecasting performance on US Coast and Danish Maritime}
\renewcommand{\arraystretch}{1.3}
\resizebox{1.45\columnwidth}{!}{
\begin{tabular}{c|cccc|cccc}
\toprule[1.6pt]

Dataset      & \multicolumn{4}{c|}{US Coast}                                                            & \multicolumn{4}{c}{Danish Maritime}                                                     \\ \hline
Timestamps       & \multicolumn{1}{c|}{1--6} & \multicolumn{1}{c|}{7--12} & \multicolumn{1}{c|}{13--24} & 1--24 & \multicolumn{1}{c|}{1--6} & \multicolumn{1}{c|}{7--12} & \multicolumn{1}{c|}{13--24} & 1--24 \\ \hline
Transformer  & \multicolumn{1}{c|}{0.0250}    & \multicolumn{1}{c|}{0.0334}     & \multicolumn{1}{c|}{0.0452}      &    0.0372  & \multicolumn{1}{c|}{0.0049}    & \multicolumn{1}{c|}{0.0078}     & \multicolumn{1}{c|}{0.0131}      &     0.0097 \\ \hline
VTP          & \multicolumn{1}{c|}{0.0166}    & \multicolumn{1}{c|}{0.0240}     & \multicolumn{1}{c|}{0.0344}      & 0.0274     & \multicolumn{1}{c|}{0.0163}    & \multicolumn{1}{c|}{0.0224}     & \multicolumn{1}{c|}{0.0316}      &  0.0255    \\ \hline
 SeqVTP  & \multicolumn{1}{c|}{0.0158}    & \multicolumn{1}{c|}{0.0230}     & \multicolumn{1}{c|}{0.0332}      &  0.0263    & 
%&&&\\ \hline
\multicolumn{1}{c|}{0.0171}    & \multicolumn{1}{c|}{0.0233}     & \multicolumn{1}{c|}{0.0325}            &   0.0263  \\ \hline
DVTP         & \multicolumn{1}{c|}{0.0139}    & \multicolumn{1}{c|}{0.0226}     & \multicolumn{1}{c|}{0.0332}      & 0.0257     & \multicolumn{1}{c|}{0.0040}    & \multicolumn{1}{c|}{0.0070}     & \multicolumn{1}{c|}{0.0125}      &   0.0090   \\ \hline
FlashFormer  & \multicolumn{1}{c|}{0.0205}    & \multicolumn{1}{c|}{0.0276}     & \multicolumn{1}{c|}{0.0387}      &   0.0314   & \multicolumn{1}{c|}{0.0058}    & \multicolumn{1}{c|}{0.0084}     & \multicolumn{1}{c|}{0.0139}      &  0.0105    \\ \hline
TimeLLM      & \multicolumn{1}{c|}{0.0065}    & \multicolumn{1}{c|}{0.0122}     & \multicolumn{1}{c|}{0.0207}      &   0.0150   & \multicolumn{1}{c|}{0.0033}    & \multicolumn{1}{c|}{0.0060}     & \multicolumn{1}{c|}{0.0107}      & 0.0077     \\ \hline
iTransformer & \multicolumn{1}{c|}{0.0085}    & \multicolumn{1}{c|}{0.0137}     & \multicolumn{1}{c|}{0.0219}      &  0.0165    & \multicolumn{1}{c|}{0.0032}    & \multicolumn{1}{c|}{0.0056}     & \multicolumn{1}{c|}{0.0103}      &   0.0073   \\ \hline
%TrAISformer  & \multicolumn{1}{c|}{}    & \multicolumn{1}{c|}{}     & \multicolumn{1}{c|}{}      &      & \multicolumn{1}{c|}{}    & \multicolumn{1}{c|}{}     & \multicolumn{1}{c|}{}      &      \\ \hline
MAKER        & \multicolumn{1}{c|}{\textbf{0.0056}}    & \multicolumn{1}{c|}{\textbf{0.0104}}     & \multicolumn{1}{c|}{\textbf{0.0182}}      &  \textbf{0.0131}    & \multicolumn{1}{c|}{\textbf{0.0028}}    & \multicolumn{1}{c|}{\textbf{0.0046}}     & \multicolumn{1}{c|}{\textbf{0.0089}}      &  \textbf{0.0063}    \\ \hline
Relative Improvement        & \multicolumn{1}{c|}{13.85\%}&\multicolumn{1}{c|}{14.75\%}&\multicolumn{1}{c|}{12.08\%}&\multicolumn{1}{c|}{12.67\%}&\multicolumn{1}{c|}{12.50\%}&\multicolumn{1}{c|}{17.86\%}&\multicolumn{1}{c|}{13.59\%}&\multicolumn{1}{c}{13.70\%}

\\ 
\toprule[1.6pt]
\end{tabular}}
\label{exp_res}
\end{table*}

\section{Experiments}
We present the experimental settings and baselines and then cover the overall findings and results of ablation studies.

\subsection{Settings}
%\subsubsection{Datasets}
We employ two quite different datasets to assess the performance of MAKER. The first dataset is sourced from the U.S. Coast Guard and is collected by onboard navigation safety devices. As the records are sampled at irregular intervals, our serialized samples also exhibit irregular intervals. Considering the speed of the vessels, we adopt a minimum sampling interval of 3 minutes. %It spans a period of 7 days and encompasses 50,961,322 records. 
The second dataset is obtained from the Danish Maritime Authority and is collected by vessels and land-based AIS stations.  Like the U.S. Coast Guard dataset, the records are sampled irregularly, resulting in serialized samples with irregular intervals in the ship trajectory data. For this particular dataset, we use a minimum sampling interval of 1 minute. %This dataset also covers 7 days but includes 108,698,248 records.

Summary statistics on the datasets are included in Table~\ref{daes}.

%\subsubsection{Implementation Details}
The experiments are conducted on a device equipped with an NVIDIA TITAN RTX 24GB GPU, and the MAKER framework is implemented in PyTorch. The code will be made publicly available upon acceptance. The batch size is set to 64, and the learning rate is set to 0.001. We use GPT-2 for contextual guidance, leveraging its pre-trained knowledge to interpret irregularly sampled trajectory points while minimizing computational costs.  We use the Adam optimizer~\cite{kingma2014adam} to update the parameters. To evaluate the performance of different MAKER configurations, we use the mean absolute error (MAE) metric for trajectory prediction. The other available features ${q}_i$ include speed over ground (SOG) and course over ground (COG). The frozen flag in Fig.\ref{fig:pip} refers to a large language model (LLM) that is used in a downstream task without any further fine-tuning.

For the Masked Sequence Encoder, the patch length $L_p$ is set to 16, the stride $S$ is set to 8, and the embedding dimensionality $d_m$ is set to 16. In the multi-modal knowledge transfer component, the hidden dimensionality $D$ of the backbone model is set to 500, and the masking ratio is set to 1/2. In the Knowledge-Based Self-Paced Learning module, we vary hyper-parameter $\lambda$ with an initial value of 0.2 and a growth rate of 1.0003.  The initial KSL value (0.2) is based on training loss, with $\lambda$ growth set at 1.0003 for balanced convergence and stability among tested rates (1.1, 1.01, 1.0001, and 1.0003).

\subsection{Experimental Results}
To evaluate the effectiveness of MAKER, we compare
with several baselines, including Transformer~\cite{han2021transformer}, VTP~\cite{tang2022model}, SeqVTP~\cite{forti2020prediction}, DVTP~\cite{capobianco2021deep}, FlashFormer~\cite{dao2022flashattention}, TimeLLM~\cite{jin2023time}, and iTransformer~\cite{liu2023itransformer}.
%\begin{itemize}
%    \item \textbf{Transformer}~\cite{han2021transformer} is utilized widely across various domains due to its powerful attention architecture and ability to capture complex patterns in sequential data.
%    \item  \textbf{VTP}~\cite{tang2022model}  utilizes a long short-term memory model to predict vessel trajectories from historical trajectory data.
%    \item \textbf{SeqVTP}~\cite{forti2020prediction}  employs an LSTM encoder-decoder architecture to forecast future trajectories.
%    \item  \textbf{DVTP}~\cite{capobianco2021deep} is a comprehensive deep-learning solution that effectively leverages an attention-based aggregation function to enable vessel trajectory prediction.
%    \item \textbf{FlashFormer}~\cite{dao2022flashattention}  computes exact attention with far fewer memory accesses. We utilize it to predict vessel trajectories.
%    \item \textbf{TimeLLM}~\cite{jin2023time} is a reprogramming framework that leverages large language models (LLMs) to enhance general time series forecasting.
%    \item \textbf{iTransformer}~\cite{liu2023itransformer} targets multivariate forecasting tasks. It applies an attention and feed-forward network on the inverted input dimensions compared to the Transformer.
    % \item \textbf{TrAISformer} is a modified transformer network that extracts long-term temporal patterns in AIS vessel trajectories in the proposed enriched space to forecast the positions of vessels several hours ahead.
%\end{itemize}

\begin{table}[]
\caption{Ablation studies of MAKER}
\centering
\resizebox{1\columnwidth}{!}{
\renewcommand{\arraystretch}{1.1}
\begin{tabular}{c|cccc|cccc}

\toprule[1.3pt]
        Dataset    & \multicolumn{4}{c|}{US Coast}                                                                     & \multicolumn{4}{c}{Danish Maritime}                                                              \\ \hline
       Timestamps     & \multicolumn{1}{c|}{1--6}     & \multicolumn{1}{c|}{7--12}   & \multicolumn{1}{c|}{13--24}  & 1--24   & \multicolumn{1}{c|}{1--6}    & \multicolumn{1}{c|}{7--12}   & \multicolumn{1}{c|}{13--24}  & 1--24    \\ \hline
MAKER-LLM   & \multicolumn{1}{c|}{0.0090}  & \multicolumn{1}{c|}{0.0147} & \multicolumn{1}{c|}{0.0242} & 0.0180 & \multicolumn{1}{c|}{0.0024} & \multicolumn{1}{c|}{0.0049} & \multicolumn{1}{c|}{0.0098} & 0.0067 \\ \hline
MAKER-de   & \multicolumn{1}{c|}{0.0051}  & \multicolumn{1}{c|}{0.0106} & \multicolumn{1}{c|}{0.0187} & 0.0133 & \multicolumn{1}{c|}{0.0022} & \multicolumn{1}{c|}{0.0050} & \multicolumn{1}{c|}{0.0097} & 0.0067 \\ \hline
MAKER-Promp & \multicolumn{1}{c|}{0.0060} & \multicolumn{1}{c|}{0.0110} & \multicolumn{1}{c|}{0.0192} & 0.0138 & \multicolumn{1}{c|}{0.0028} & \multicolumn{1}{c|}{0.0047} & \multicolumn{1}{c|}{0.0091} & 0.0065 \\ \hline
MAKER-KSL   & \multicolumn{1}{c|}{0.0059}        & \multicolumn{1}{c|}{0.0106}       & \multicolumn{1}{c|}{0.0182}       &   0.0132     & \multicolumn{1}{c|}{0.0027} & \multicolumn{1}{c|}{0.0047} & \multicolumn{1}{c|}{0.0090} & 0.0064 \\ \hline
MAKER-MKT   & \multicolumn{1}{c|}{0.0248} & \multicolumn{1}{c|}{0.0334} & \multicolumn{1}{c|}{0.0451} & 0.0371 & \multicolumn{1}{c|}{0.0241} & \multicolumn{1}{c|}{0.0320} & \multicolumn{1}{c|}{0.0431} & 0.0356 \\ \hline
MAKER       & \multicolumn{1}{c|}{0.0056}  & \multicolumn{1}{c|}{0.0104} & \multicolumn{1}{c|}{0.0182} & 0.0131 & \multicolumn{1}{c|}{0.0028} & \multicolumn{1}{c|}{0.0046} & \multicolumn{1}{c|}{0.0089} & 0.0063 \\  \toprule[1.3pt]
\end{tabular}}

\end{table}

Table~\ref{exp_res} reports the experimental findings for MAKER and the baselines on the two datasets. The best results are highlighted in bold. (1)~The most competitive methods,  TimeLLM and iTransformer, perform well, particularly on the Danish Maritime dataset. They achieve good results in predicting the next 6 timestamps. However, as the prediction horizon is increased, their performance declines sharply. (2)~The US Coast dataset, characterized by a minimum sampling interval of 3 minutes, highlights the increased difficulty in achieving high performance compared to the Danish Maritime dataset. (3)~While MAKER demonstrates significantly better performance primarily in predicting future 7--12 timestamps, the overall experimental results highlight its strengths across all time horizons.

\begin{table}[h!]

\caption{Comparison of complexity and irregularity}
\centering

\renewcommand{\arraystretch}{1.05}

\resizebox{1\columnwidth}{!}{
\begin{tabular}{l|c|c|c|c|c|c}
\toprule[1.2pt]
{Danish Maritime}         & \multicolumn{3}{c|}{{Spatial Complexity}} & \multicolumn{3}{c}{{Time Irregularity}} \\ \hline
{Level}      & Low        & Medium     & High        & Low        & Medium     & High        \\ \hline
{MAKER}      & 0.0002     & 0.0048     & 0.0156      & 0.0058     & 0.0050     & 0.0094      \\ \hline
{MAKER-LLM}  & 0.0002-    & 0.0050↓    & 0.0166↓     & 0.0062↓    & 0.0054↓    & 0.0099↓     \\ \hline
{MAKER-KSL}  & 0.0001↑    & 0.0048-    & 0.0158↓     & 0.0058-    & 0.0050-    & 0.0096↓     \\ \hline
{TimeLLM}    & 0.0004↓    & 0.0061↓    & 0.0180↓     & 0.0071↓    & 0.0063↓    & 0.0110↓     \\ \hline
{iTransformer} & 0.0001↑   & 0.0053↓    & 0.0188↓     & 0.0064↓    & 0.0059↓    & 0.0112↓     \\ 
\toprule[1.2pt]
\end{tabular}}

\label{tab:c}

\end{table}

\subsection{Ablation Studies}
\label{abstud}
To explore the effectiveness of each component of MAKER, we design ablation studies to compare different variants of MAKER. (1) MAKER-LLM removes the LLM-guided sequence model to verify the effectiveness of the large language module. (2) MAKER-de removes the decoder to
verify the role of the iTransformer decoder. (3) MAKER-Prompt removes
the prompt hint information. (4) MAKER-KSL removes the knowledge-based self-paced learning module. (5) MAKER-MKT removes the multi-modal knowledge transfer component, which means the module does not use the transfer component to bridge the information between the trajectory records and textual information.

The ablation studies are conducted on two datasets, and we use the same forecasting lengths as in the previous experiments. We can observe that: (1) The best results are achieved by MAKER for predicting future 7--12 timestamps, 13--24 timestamps, and 1--24 timestamps, demonstrating the effectiveness of each component in vessel trajectory prediction. (2) The performance of MAKER-MKT is considerably lower on both datasets, indicating that the multi-modal knowledge transfer component is crucial for bridging the gap between sequence and textual models. (3) The performance of MAKER-LLM, which excludes the large language model, is also reduced, highlighting the importance of the large language model as a comprehensive pre-trained model.

To assess the model's ability to handle spatial complexity and temporal irregularity, we divide and categorize trajectory complexity and temporal irregularity into three levels based on quartiles:
scores above the upper quartile (75th percentile) are classified as ``High, '' and scores below the lower
quartile (25th percentile) are classified as ``Low.'' The remaining scores are classified between the quartiles as ``Medium.'' Spatial complexity is measured using the standard deviation of Euclidean distances between consecutive points, with higher values indicating more complexity. To measure temporal irregularity, we divide the sequence of time intervals into
the first 24 steps (input) and the last 24 steps (prediction). We calculate the standard deviation for
each part, using the relative change between these values as an indicator of irregularity.
We evaluate the KSL module's predictive accuracy and the LLM’s handling of irregular data across these categorized levels, with results on the Danish Maritime dataset presented in Table~\ref{tab:c}. MAKER-LLM excludes the LLM-guided sequence model, while MAKER-KSL excludes knowledge-based self-paced learning. We compare with the two most competitive baselines, TimeLLM and iTransformer.

The results in the table yield two key findings. (1)~MAKER's advantage over MAKER-LLM becomes more evident as spatial complexity and temporal irregularity increase, highlighting the LLM model's effectiveness in handling these challenges. While MAKER shows a slight performance decrease on simpler data compared to MAKER-KSL, its performance improves significantly with increased complexity and irregularity. This is evidence that the proposed method is suitable for complex trajectory patterns.
(2)~Compared to iTransformer and TimeLLM, MAKER’s advantage becomes more pronounced as trajectory complexity and temporal irregularity increase. In contrast to MAKER-LLM and MAKER-KSL, MAKER shows a notably larger performance improvement over iTransformer and TimeLLM, highlighting its effectiveness in managing complexity and irregularity. These findings indicate that MAKER has strong generalization capabilities, especially in more challenging environments.

\vspace{5pt}

\section{Conclusion}
We propose a multi-modal knowledge-enhanced framework~(\textsf{MAKER}) for vessel trajectory prediction. The framework can discern complex trajectory patterns despite irregular sampling intervals by exploiting multi-modal trajectory information. \textsf{MAKER} contains two key modules, including a large language model-guided knowledge transfer~(LKT) module and a knowledge-based self-paced learning~(KSL) module.
The LKT module utilizes multi-modal data to ease the impact of irregular sampling intervals. The KSL module progressively incorporates complex patterns into the training process to enhance the model's ability to process complex trajectory data.  
The experimental results confirm the effectiveness of the two modules.

In future research, it is of interest to explore additional types of kinematic knowledge to improve trajectory prediction and extend the multimodal knowledge-enhancement framework to other maritime scenarios, such as collision warnings.
 
%There are some topics we will study in the future.
%It is meaningful to further explore the kinematic knowledge that is more helpful for trajectory prediction tasks. Then, it is also meaningful to extend the multimodal knowledge enhancement framework to other maritime scenarios (e.g., collision warning).

\bibliographystyle{acm}

\bibliography{mybib}

\begin{thebibliography}{10}

\bibitem{alizadeh2021prediction}
{\sc Alizadeh, D., Alesheikh, A.~A., and Sharif, M.}
\newblock Prediction of vessels locations and maritime traffic using similarity measurement of trajectory.
\newblock {\em Annals of GIS 27}, 2 (2021), 151--162.

\bibitem{bengio2009curriculum}
{\sc Bengio, Y., Louradour, J., Collobert, R., and Weston, J.}
\newblock Curriculum learning.
\newblock {\em International Conference on Machine Learning\/} (2009), 41--48.

\bibitem{capobianco2021deep}
{\sc Capobianco, S., Millefiori, L.~M., Forti, N., Braca, P., and Willett, P.}
\newblock Deep learning methods for vessel trajectory prediction based on recurrent neural networks.
\newblock {\em IEEE Transactions on Aerospace and Electronic Systems 57}, 6 (2021), 4329--4346.

\bibitem{cervera2011satellite}
{\sc Cervera, M.~A., Ginesi, A., and Eckstein, K.}
\newblock Satellite-based vessel automatic identification system: A feasibility and performance analysis.
\newblock {\em International Journal of Satellite Communications and Networking 29}, 2 (2011), 117--142.

\bibitem{dao2022flashattention}
{\sc Dao, T., Fu, D., Ermon, S., Rudra, A., and R{\'e}, C.}
\newblock Flashattention: Fast and memory-efficient exact attention with io-awareness.
\newblock {\em Advances in Neural Information Processing Systems 35\/} (2022), 16344--16359.

\bibitem{forti2020prediction}
{\sc Forti, N., Millefiori, L.~M., Braca, P., and Willett, P.}
\newblock Prediction of vessel trajectories from {AIS} data via sequence-to-sequence recurrent neural networks.
\newblock {\em IEEE International Conference on Acoustics, Speech and Signal Processing\/} (2020), 8936--8940.

\bibitem{gruver2024large}
{\sc Gruver, N., Finzi, M., Qiu, S., and Wilson, A.~G.}
\newblock Large language models are zero-shot time series forecasters.
\newblock {\em Advances in Neural Information Processing Systems 36\/} (2024).

\bibitem{han2021transformer}
{\sc Han, K., Xiao, A., Wu, E., Guo, J., Xu, C., and Wang, Y.}
\newblock Transformer in transformer.
\newblock {\em Advances in Neural Information Processing Systems 34\/} (2021), 15908--15919.

\bibitem{jin2023time}
{\sc Jin, M., Wang, S., Ma, L., Chu, Z., Zhang, J.~Y., Shi, X., Chen, P.-Y., Liang, Y., Li, Y.-F., Pan, S., et~al.}
\newblock {Time-LLM}: Time series forecasting by reprogramming large language models.
\newblock {\em arXiv preprint arXiv:2310.01728\/} (2023).

\bibitem{kingma2014adam}
{\sc Kingma, D.~P., and Ba, J.~L.}
\newblock Adam: a method for stochastic optimization.
\newblock {\em arXiv preprint arXiv:1412.6980\/} (2014).

\bibitem{koccak2021comparative}
{\sc Ko{\c{c}}ak, S.~T., and Yercan, F.}
\newblock Comparative cost-effectiveness analysis of arctic and international shipping routes: A fuzzy analytic hierarchy process.
\newblock {\em Transport Policy 114\/} (2021), 147--164.

\bibitem{kumar2010self}
{\sc Kumar, M., Packer, B., and Koller, D.}
\newblock Self-paced learning for latent variable models.
\newblock {\em Advances in Neural Information Processing Systems 23\/} (2010).

\bibitem{lee2023learning}
{\sc Lee, S., Park, T., and Lee, K.}
\newblock Learning to embed time series patches independently.
\newblock {\em arXiv preprint arXiv:2312.16427\/} (2023).

\bibitem{li2023self}
{\sc Li, Y., Wu, K., and Liu, J.}
\newblock Self-paced {ARIMA} for robust time series prediction.
\newblock {\em Knowledge-Based Systems 269\/} (2023), 110489.

\bibitem{liu2023itransformer}
{\sc Liu, Y., Hu, T., Zhang, H., Wu, H., Wang, S., Ma, L., and Long, M.}
\newblock {iTransformer}: Inverted transformers are effective for time series forecasting.
\newblock {\em arXiv preprint arXiv:2310.06625\/} (2023).

\bibitem{mazzarella2015knowledge}
{\sc Mazzarella, F., Arguedas, V.~F., and Vespe, M.}
\newblock Knowledge-based vessel position prediction using historical {AIS} data.
\newblock {\em Sensor Data Fusion: Trends, Solutions, Applications\/} (2015), 1--6.

\bibitem{qi2016trajectory}
{\sc Qi, L., and Zheng, Z.}
\newblock Trajectory prediction of vessels based on data mining and machine learning.
\newblock {\em Journal of Digital Information Management 14}, 1 (2016), 33--40.

\bibitem{raamadhurai2019curio}
{\sc Raamadhurai, S., Baker, R., and Poduval, V.}
\newblock Curio smartchat: a system for natural language question answering for self-paced k-12 learning.
\newblock {\em Workshop on Innovative Use of NLP for Building Educational Applications\/} (2019), 336--342.

\bibitem{tang2022model}
{\sc Tang, H., Yin, Y., and Shen, H.}
\newblock A model for vessel trajectory prediction based on long short-term memory neural network.
\newblock {\em Journal of Marine Engineering \& Technology 21}, 3 (2022), 136--145.

\bibitem{tang2019tensor}
{\sc Tang, Y., Xie, Y., Yang, X., Niu, J., and Zhang, W.}
\newblock Tensor multi-elastic kernel self-paced learning for time series clustering.
\newblock {\em IEEE Transactions on Knowledge and Data Engineering 33}, 3 (2019), 1223--1237.

\bibitem{vaswani2017attention}
{\sc Vaswani, A., Shazeer, N., Parmar, N., Uszkoreit, J., Jones, L., Gomez, A.~N., Kaiser, {\L}., and Polosukhin, I.}
\newblock Attention is all you need.
\newblock {\em arXiv preprint arXiv:1706.03762\/} (2017).

\bibitem{wan2020self}
{\sc Wan, Y., Yang, B., Wong, D.~F., Zhou, Y., Chao, L.~S., Zhang, H., and Chen, B.}
\newblock Self-paced learning for neural machine translation.
\newblock {\em arXiv preprint arXiv:2010.04505\/} (2020).

\bibitem{wang2022fedspl}
{\sc Wang, Q., and Zhou, Y.}
\newblock {FedSPL}: federated self-paced learning for privacy-preserving disease diagnosis.
\newblock {\em Briefings in Bioinformatics 23}, 1 (2022), bbab498.

\bibitem{wang2020adaptive}
{\sc Wang, Q., Zhou, Y., Zhang, W., Tang, Z., and Chen, X.}
\newblock Adaptive sampling using self-paced learning for imbalanced cancer data pre-diagnosis.
\newblock {\em Expert Systems with Applications 152\/} (2020), 113334.

\bibitem{wu2017semi}
{\sc Wu, S., Ji, Q., Wang, S., Wong, H.-S., Yu, Z., and Xu, Y.}
\newblock Semi-supervised image classification with self-paced cross-task networks.
\newblock {\em IEEE Transactions on Multimedia 20}, 4 (2017), 851--865.

\bibitem{xu2018multi}
{\sc Xu, W., Liu, W., Huang, X., Yang, J., and Qiu, S.}
\newblock Multi-modal self-paced learning for image classification.
\newblock {\em Neurocomputing 309\/} (2018), 134--144.

\bibitem{yang2024self}
{\sc Yang, S., Deng, X., and Song, D.}
\newblock Self-paced learning long short-term memory based on intelligent optimization for robust wind power prediction.
\newblock {\em IET Control Theory \& Applications\/} (2024).

\bibitem{yu2023cgf}
{\sc Yu, H., Hu, J., Zhou, X., Guo, C., Yang, B., and Li, Q.}
\newblock {CGF}: A category guidance based {PM2.5} sequence forecasting training framework.
\newblock {\em IEEE Transactions on Knowledge and Data Engineering 35}, 10 (2023), 10125--10139.

\bibitem{yu2023temporal}
{\sc Yu, X., Chen, Z., Ling, Y., Dong, S., Liu, Z., and Lu, Y.}
\newblock Temporal data meets {LLM}--explainable financial time series forecasting.
\newblock {\em arXiv preprint arXiv:2306.11025\/} (2023).

\end{thebibliography}

\balance

\end{document}